\documentclass[unnumsec,webpdf,contemporary,large]{oup-authoring-template}

\graphicspath{{Fig/}}

\theoremstyle{thmstyleone}%
\theoremstyle{thmstyletwo}%
\theoremstyle{thmstylethree}%

\usepackage{tabularx}
\usepackage{booktabs}
\usepackage{flushend}
\usepackage{microtype}
\begin{document}

\journaltitle{Journal Title Here}
\DOI{DOI HERE}
\copyrightyear{2022}
\pubyear{2019}
\access{Advance Access Publication Date: Day Month Year}
\appnotes{Paper}

\firstpage{1}


\title[Short Article Title]{PGCLODA: Prompt-Guided Graph Contrastive Learning for Oligopeptide-Infectious Disease Association Prediction}

\author[1,2]{Dayu Tan\ORCID{0000-0002-3960-8852}}
\author[2,3]{Jing Chen}
\author[2,3]{Xiaoping Zhou}
\author[2,3,$\ast$]{Yansen Su}
\author[2,3]{Chunhou Zheng}

\authormark{Author Name et al.}

\address[1]{\orgdiv{Institute of Physical Science and Information Technology}, \orgname{Anhui University}, \orgaddress{\street{Hefei}, \postcode{230601}, \state{Anhui}, \country{China}}}

\address[2]{\orgdiv{Key Laboratory of Intelligent Computing and Signal Processing of Ministry of Education}, \orgname{Anhui University}, \orgaddress{\street{Hefei}, \postcode{230601}, \state{Anhui}, \country{China}}}

\address[3]{\orgdiv{School of Artificial Intelligence}, \orgname{Anhui University}, \orgaddress{\street{Hefei}, \postcode{230601}, \state{Anhui}, \country{China}}}


\corresp[$\ast$]{Corresponding author. \href{email:email-id.com}{suyansen@ahu.edu.cn}}

\received{Date}{0}{Year}
\revised{Date}{0}{Year}
\accepted{Date}{0}{Year}



\abstract{Infectious diseases continue to pose a serious threat to public health, underscoring the urgent need for effective computational approaches to screen novel anti-infective agents. Oligopeptides have emerged as promising candidates in antimicrobial research due to their structural simplicity, high bioavailability, and low susceptibility to resistance. Despite their potential, computational models specifically designed to predict associations between oligopeptides and infectious diseases remain scarce. This study introduces a prompt-guided graph-based contrastive learning framework (PGCLODA) to uncover potential associations. A tripartite graph is constructed with oligopeptides, microbes, and diseases as nodes, incorporating both structural and semantic information. To preserve critical regions during contrastive learning, a prompt-guided graph augmentation strategy is employed to generate meaningful paired views. A dual encoder architecture, integrating Graph Convolutional Network (GCN) and Transformer, is used to jointly capture local and global features. The fused embeddings are subsequently input into a multilayer perceptron (MLP) classifier for final prediction. Experimental results on a benchmark dataset indicate that PGCLODA consistently outperforms state-of-the-art models in AUROC, AUPRC, and accuracy. Ablation and hyperparameter studies confirm the contribution of each module. Case studies further validate the generalization ability of PGCLODA and its potential to uncover novel, biologically relevant associations. These findings offer valuable insights for mechanism-driven discovery and oligopeptide-based drug development. The source code of PGCLODA is available online at https://github.com/jjnlcode/PGCLODA.}
\keywords{Oligopeptides, Infectious disease, Contrastive learning, Prompt learning, Association prediction}


\maketitle

\section{I Introduction}
Infectious diseases remain a major global public health threat. The emergence of drug-resistant bacterial strains and novel pathogens poses significant challenges to current anti-infective therapies~\citep*{salam2023antimicrobial,zhu2025recent}. Although widely used in clinical settings, traditional small-molecule antibiotics are prone to inducing multidrug resistance~\citep*{bergkessel2023small,liu2019antibiotic}. Their broad-spectrum activity may also disrupt commensal microbial communities, leading to secondary infections~\citep*{rea2011effect}. Peptide-based therapeutics, composed of medium-length amino acid chains, offer a lower risk of resistance induction compared to small molecules. These peptides exert antimicrobial effects through multiple mechanisms, such as disrupting bacterial membranes, inhibiting virulence factors, and modulating host immune responses~\citep*{duarte2023antimicrobial,koprivnjak2011bacterial}. However, their clinical translation is hindered by challenges including synthetic complexity, limited in vivo stability, and suboptimal structural properties. These limitations underscore the need to develop alternative therapeutic agents with potent efficacy, improved resistance profiles, and enhanced pharmacological stability.

Oligopeptides, consisting of 2 to 9 amino acid residues, have garnered increasing interest in antimicrobial drug research owing to their structural simplicity, efficient synthesis, modifiability, and inherent stability~\citep*{apostolopoulos2021global,molchanova2017advances}. Compared with conventional peptide drugs, oligopeptides exhibit lower molecular weight, enhanced bioavailability, and superior membrane permeability~\citep*{verma2021challenges,han2019multifunctional}. Moreover, they can be rationally engineered to improve target affinity and functional performance, thereby offering greater design flexibility and drug development potential. In recent years, numerous studies have experimentally validated the antimicrobial efficacy of oligopeptides against pathogenic microorganisms.~\citep*{li2024artificial,fong2024critical} For instance, Wang et al.~\citep*{wang2018oligopeptide} computationally designed a pentapeptide, LPRDA, which specifically inhibits Sortase A—an enzyme in Staphylococcus aureus—thereby reducing bacterial adhesion and invasion. The pentapeptide demonstrated strong antimicrobial activity in a murine mastitis model. Similarly, Lu et al.~\citep*{lu2024antibacterial} isolated a natural oligopeptide, X33 AMOP, from \textit{Streptomyces lavendulae}, which exhibited potent antibacterial effects against multidrug-resistant \textit{Acinetobacter baumannii}. The reported mechanisms of action included membrane disruption, induction of oxidative stress, and interference with energy metabolism. Furthermore, Silva et al.~\citep*{silva2016anti} introduced a short oligopeptide segment, FLPII, into the natural antimicrobial peptide Clavanin A, thereby constructing a synthetic derivative, Clavanin-MO, with significantly enhanced antimicrobial and immunomodulatory functions. Collectively, these studies highlight that oligopeptides not only possess intrinsic antimicrobial properties but also serve as functional scaffolds in peptide drug design, exhibiting substantial potential to enhance therapeutic efficacy and biological stability.

Although oligopeptides exhibit significant potential in combating infectious diseases~\citep*{mhlongo2023antimicrobial}, computational models capable of systematically predicting their associations remain scarce. Most existing studies have concentrated on experimentally validating specific oligopeptide sequences or elucidating their biological mechanisms~\citep*{goles2024peptide,cryptic2022genome}, yet they lack generalizable and scalable models for association prediction. In bioinformatics, considerable efforts have been devoted to predicting molecular associations, including miRNA–disease~\citep*{peng2023mhclmda,liu2022identification,chen2012rwrmda}, circRNA–disease~\citep*{ma2021crpgcn,peng2023predicting}, small molecule–disease \citep*{ding2021machine}, and microbe–disease~\citep*{long2021predicting} associations. These approaches encompass network-based path propagation, feature-based matrix factorization, ensemble learning algorithms, and graph neural networks (GNNs), which have gained increasing popularity in recent years. Although these methods have achieved considerable success in binary association prediction tasks, they fall short in modeling the complex ternary interaction pathways commonly present in oligopeptide–disease relationships. On one hand, oligopeptides frequently influence disease progression by modulating specific microbes, thereby forming multi-hop paths such as "oligopeptide–microbe–disease". Traditional binary models fail to explicitly capture such heterogeneous and compositional dependencies, often resulting in the omission of critical relational information. On the other hand, although certain existing models distinguish between node types, they often overlook the semantic roles and directional dependencies inherent to ternary structures, thereby limiting their ability to represent the regulatory logic of oligopeptides in microbial modulation and disease progression. Therefore, there is an urgent need for representation learning frameworks capable of modeling multi-entity and multi-relation structures, thereby enabling accurate and interpretable prediction of oligopeptide–disease associations.

To tackle these challenges, this study introduces Prompt-Guided Graph Contrastive Learning for Oligopeptide–Disease Association Prediction (PGCLODA), a heterogeneous graph-based framework leveraging contrastive learning to predict oligopeptide–infectious disease associations. PGCLODA models oligopeptides, microbes, and diseases as three distinct types of nodes, constructing a ternary heterogeneous graph. The resulting graph integrates both structural and semantic information derived from multiple relational sources. Building upon this graph, a prompt-guided augmentation mechanism is developed to generate positive and negative graph pairs for contrastive learning. The augmentation mechanism preserves the structural integrity of representative oligopeptide nodes while perturbing edges in the surrounding regions. This design enhances PGCLODA’s ability to identify subtle local structural variations. Subsequently, a dual-encoder architecture that combines Graph Convolutional Networks (GCNs) and Transformers is employed to capture both local connectivity patterns and global semantic dependencies, thereby enabling hierarchical feature embeddings. The final embeddings of oligopeptide and disease nodes are concatenated and passed through a multilayer perceptron (MLP) to predict potential associations. Compared with existing methods, PGCLODA effectively handles multi-hop paths and heterogeneous node types. Moreover, the prompt-guided strategy enhances the preservation of critical graph structures during augmentation, whereas contrastive learning strengthens the discriminative capacity of embeddings and improves overall predictive performance.
\section{II Related work}\label{relatedwork}
Predicting potential associations between biomolecules has become a vital tool for elucidating disease mechanisms and identifying candidate therapeutic targets, garnering increasing attention in areas such as miRNA–disease, circRNA–disease, and microbe–disease association prediction. Earlier studies primarily relied on classical methods, including network topology-based propagation and matrix factorization.~\citep*{heine2016survey,hensel2021survey} In 2012, Chen et al.~\citep*{chen2012rwrmda} introduced RWRMDA (Random Walk with Restart for MiRNA–Disease Association), a method that propagates similarity scores across the miRNA–disease network via a random walk mechanism. This approach facilitates the effective inference of previously unknown associations. In 2018, Li et al.~\citep*{cui2019l2} proposed the GRMF (Graph Regularized Matrix Factorization) model, which integrates graph-based regularization into low-rank factorization of the association matrix to preserve local similarity structures. This strategy enhances the predictive performance by preserving semantic relationships in the latent space. Although these methods perform well on dense networks and provide strong interpretability, they often struggle to capture the complex semantic and structural interactions among heterogeneous node types.

With the advancement of graph neural networks (GNNs), an increasing number of studies have employed graph-based representation learning to uncover potential associations among biomolecules. In 2021, Lai et al.~\citep*{tang2021multi} introduced MMGCN (Multi-view Multichannel Graph Convolutional Network), which integrates multiple similarity perspectives of miRNAs and diseases via multi-view GCN and channel-wise convolutional fusion. This design enhances the discriminative capability of node representations. Also in 2021, Ma et al.~\citep*{ma2021crpgcn} introduced CRPGCN (CircRNA-Disease Prediction via Graph Convolutional Network and Random Walk), which combines attribute features and graph structures of circRNAs and diseases.The model further incorporates walk-based features and semantic representations to enhance structural expressiveness. These methods provide valuable attempts to incorporate graph structural information; however, most of them are built on static homogeneous graphs and lack explicit modeling of node types and multi-hop semantic paths. To achieve structure-aware modeling, attention mechanisms have been extensively adopted to improve the expressiveness of node interactions. In 2023, Li et al.~\citep*{peng2023predicting} proposed GATCL2CD (Graph Attention and Contrastive Learning for CircRNA–Disease Association), which integrates multi-head graph attention with contrastive learning to enhance sensitivity to local structural differences among nodes. In 2024, Huang et al.~\citep*{huang2024hierarchical} introduced HDGAT (Hierarchical Dual-level Graph Attention Network), which jointly models drug–disease associations by employing both global and local attention mechanisms. This design enhances the ability to identify key semantic edges within the graph. These models demonstrate strong expressive capacity in structural selection and edge weight modeling, effectively capturing local structural differences. However, they primarily focus on neighbor-level interactions and lack holistic modeling of global structures and upstream–downstream semantic dependencies.

In recent years, contrastive learning~\citep{tan2024scamac,su2023denoising} and graph augmentation have attracted growing attention in biological graph representation learning. Zhao et al.~\citep*{zhao2024ognnmda} proposed OGNNMDA (Over-smoothing-aware Graph Neural Network with Contrastive Learning for miRNA–Disease Association Prediction), which integrates graph perturbation strategies into a contrastive learning framework. Through the introduction of a contrastive view generation module, OGNNMDA effectively mitigates the over-smoothing problem and enhances both the robustness and structural discriminability of node representations.In the same year, He et al.~\citep{he2024fusing} proposed DRGBCN (Dual Representation and Global-Contextual Contrastive Network), which constructs both global and semantic graphs and imposes cross-view contrastive constraints to guide the learning of unified and highly discriminative node representations. Both OGNNMDA and DRGBCN primarily focus on enhancing robustness against local perturbations and optimizing semantic consistency. However, these methods fail to explicitly capture cross-path dependencies among multiple node types or to represent “regulation–transmission–action” ternary structures within heterogeneous graphs. In addition, several studies have focused on modeling ternary interaction structures. In 2023, Liu et al.~\citep*{liu2023hgnnlda} introduced HGNNLDA (Heterogeneous Graph Neural Network for LncRNA–Disease Association), which constructs a unified heterogeneous graph incorporating lncRNAs, miRNAs, and diseases. The model employs restart random walks to sample neighbors and applies heterogeneous attention mechanisms to aggregate information across diverse node types, significantly improving prediction performance. However, HGNNLDA remains limited in modeling complex ternary paths and cross-entity interactions, as it primarily focuses on attention allocation between neighboring nodes and lacks explicit representation of hierarchical path semantics and functional logic across entity types. In 2025, Kang et al.~\citep*{kang2025comprehensive} proposed TriMoGCL, a graph contrastive learning framework tailored for triplet motif classification in heterogeneous biomedical graphs. The framework defines seven representative structural motifs and employs both node-level and prototype-level contrastive learning to enhance semantic discrimination. However, TriMoGCL relies on predefined motif templates and lacks a flexible mechanism to capture diverse and task-specific semantic dependencies.

In summary, recent studies have achieved significant progress in representation learning, encompassing multi-view feature integration, attention mechanism optimization, graph structure enhancement, and contrastive learning strategies. Nevertheless, current methods remain limited when applied to complex heterogeneous graphs characterized by multiple node types and multi-hop semantic dependencies. A representative case is the oligopeptide–microbe–disease paradigm, in which existing approaches struggle with structural expressiveness, lack explicit modeling of semantic dependency chains, and exhibit limited robustness. Therefore, developing a unified graph representation learning framework capable of jointly modeling heterogeneous entities, semantic path dependencies, and structure-aware enhancement mechanisms is essential. Such a framework is expected to enhance both the predictive accuracy and interpretability of potential biomolecular associations.

\section{III Method}\label{method}
\begin{figure*}[htbp]
  \centering
  \includegraphics[width=\textwidth]{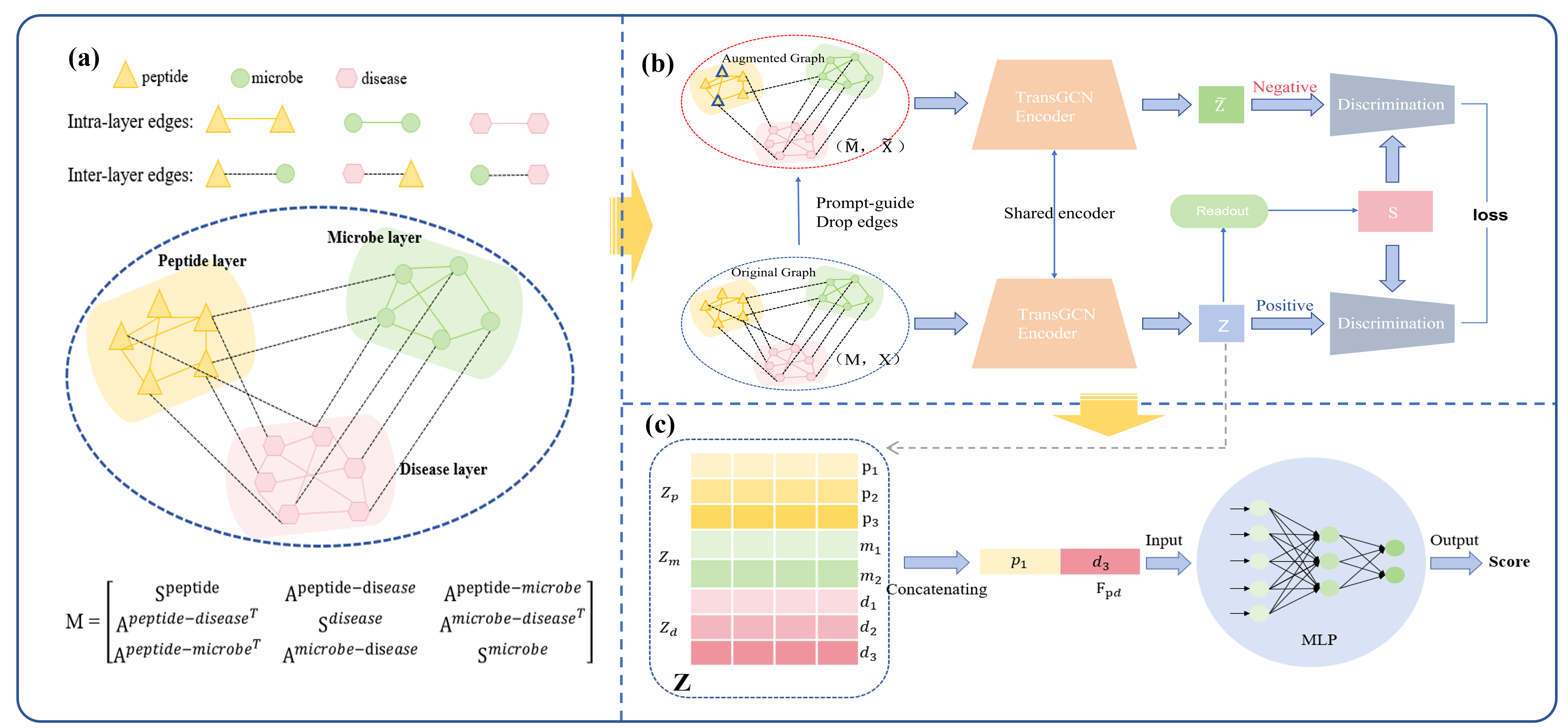}
  \caption{Overview of the PGCLODA framework comprising three core components: (1) Ternary heterogeneous graph construction with oligopeptides, microbes, and diseases; (2) Prompt-guided graph augmentation and dual-encoder embedding extraction for generating contrastive views; (3) Association prediction via contrastive learning and a multilayer perceptron (MLP) classifier.}
  \label{fig:PGCLODA}
 \end{figure*}

This study introduces a heterogeneous graph-based contrastive learning framework designed to predict potential associations between oligopeptides and infectious diseases. FThe overall architecture of the proposed framework is depicted in Fig.~\ref{fig:PGCLODA}, comprising four main components: heterogeneous graph construction, prompt-guided graph augmentation, dual-encoder embedding learning, and contrastive learning optimization. The framework is constructed upon a ternary heterogeneous graph that integrates oligopeptides, microbes, and diseases as distinct node types, while embedding both similarity and association information within a unified representation. A prompt-aware selection mechanism is employed to identify representative nodes, referred to as prompt nodes. Edges connecting prompt nodes to non-prompt nodes are then randomly perturbed to generate an augmented view of the original graph for contrastive training. Both the original and augmented graphs are processed by a dual-encoder module, which combines a Graph Convolutional Network (GCN) and a Transformer to respectively capture local structural features and global semantic dependencies, thereby refining node embeddings. For each node, embeddings derived from both the original and augmented graphs are paired with the globally pooled representation of the original graph, forming positive and negative sample pairs for contrastive discrimination. Finally, the refined embeddings of oligopeptide and disease nodes are concatenated and input into a multilayer perceptron (MLP) classifier to predict their potential associations.

 \subsection{A. Data Preprocessing and Similarity Computation}\label{date}
 To construct a high-quality heterogeneous graph, the present study systematically collected association data among oligopeptides, microbes, and diseases, followed by standardized preprocessing and filtering. Experimentally validated oligopeptide sequences were primarily retrieved from public repositories, especially DBAASP (Database of Antimicrobial Activity and Structure of Peptides)~\citep*{pirtskhalava2021dbaasp,gogoladze2014dbaasp}, which contains a wide range of natural and synthetic antimicrobial peptides. Only peptides with sequence lengths between 2 and 9 amino acids were retained, and CD-Hit~\citep*{li2006cd,fu2012cd} was applied with a similarity threshold of 0.7 to eliminate redundancy. Microbial entities were standardized based on international nomenclature guidelines, such as merging strain-level identifiers (e.g., Staphylococcus aureus ATCC 29213, BAA-44, and ATCC 43300) into their species-level representation (e.g., Staphylococcus aureus). Microbe-disease associations were subsequently obtained from the Disbiome database~\citep*{janssens2018disbiome}. These data were further integrated to infer oligopeptide–disease associations, which were subsequently used for graph construction. The statistics of node and edge types in the constructed heterogeneous graph are summarized in Table 1.

\begin{table}[htbp]
  \centering
  \caption{Statistics of nodes and edges in the constructed heterogeneous graph.}
  \label{tab:node-edge}
  \begin{tabular}{l r l r}
    \toprule
    \textbf{Node} & \textbf{Num} & \textbf{Edge} & \textbf{Num} \\
    \midrule
    Peptide & 1084 & Peptide--microbe & 1130 \\
    Microbe & 81   & Microbe--disease & 544 \\
    Disease & 173  & Peptide--disease & 14643 \\
    -- & -- & Peptide--peptide & 1175056 \\
    -- & -- & Microbe--microbe & 6561 \\
    -- & -- & Disease--disease & 29929 \\
    \bottomrule
  \end{tabular}
\end{table}

 Following data preprocessing, association and similarity matrices among oligopeptides, microbes, and diseases were constructed to enhance the semantic expressiveness of the heterogeneous graph. In this study, an association is defined as a direct connection between two nodes, indicating a biologically functional relationship or interaction. For instance, an oligopeptide–microbe association reflects the antimicrobial effect of the oligopeptide; a microbe–disease association indicates the role of microbial infection in disease onset; and an oligopeptide–disease association denotes the therapeutic or interventional potential of the oligopeptide in treating a given disease. To encode such associations, an association matrix \(A\)  was defined, where each element \( A_{ij} \) indicates the presence or absence of a relationship between nodes \( i \) and node \( j \):
 \begin{equation}
 A_{ij} =
 \begin{cases}
 1, & \text{if an association exists between nodes } i \text{ and } j \\
 0, & \text{otherwise}
 \end{cases}.
 \end{equation}
 
 In addition, similarity matrices were separately constructed for oligopeptides, microbes, and diseases to enhance both structural and semantic connectivity among homogeneous nodes. The similarity between oligopeptides was calculated based on the Smith–Waterman~\citep*{pearson1991searching} local sequence alignment algorithm. Given two oligopeptide sequences \( p_{i} \) and \( p_{j} \), their similarity score is defined as:
 \begin{equation}
 S_p(i, j) = SW(p_i, p_j),
 \end{equation} where \( SW(p_i, p_j) \) 
 denotes the local alignment score computed using the Smith-Waterman algorithm. This algorithm assigns positive scores for matches and imposes penalties for mismatches. Gap penalties are incorporated during the alignment process based on predefined opening and extension costs. The final similarity score corresponds to the highest alignment score among all possible local alignment paths.

 Microbe–microbe and disease–disease similarities were computed using the Gaussian Interaction Profile (GIP) kernel~\citep*{van2011gaussian}. The GIP kernel measures the similarity between entities in interaction space based on their interaction profiles with associated entities. The corresponding formulations are as follows:
 \begin{equation}
 S_m(m_i, m_j) = \exp\left(-\gamma_m \cdot \left\| G(m_i) - G(m_j) \right\|^2 \right),
 \end{equation}
 \begin{equation}
 S_d(d_i, d_j) = \exp\left(-\gamma_d \cdot \left\| G(d_i) - G(d_j) \right\|^2 \right).
 \end{equation}

 Herein, \( G(m_i) \) denotes the binary interaction profile of microbe \( m_i \) with all diseases, and \( G(d_i) \) represents the interaction profile of disease \( d_i \) with all microbes. The parameters \( \gamma_m \) and \( \gamma_d \) represent the bandwidths of the Gaussian kernels for microbes and diseases, respectively. To ensure scale consistency across entities, the bandwidth parameters \( \gamma_m \) and \( \gamma_d \) were normalized as follows:
 \begin{equation}
 \gamma_m = \gamma_m' \cdot \frac{1}{n_m} \sum_{i=1}^{n_m} \left\| G(m_i) \right\|,
 \end{equation}

 \begin{equation}
 \gamma_d = \gamma_d' \cdot \frac{1}{n_d} \sum_{i=1}^{n_d} \left\| G(d_i) \right\|.
 \end{equation}
 
 In Eqs. (5) and (6), \( n_m \) and \( n_d \) denote the number of microbe and disease nodes, respectively. The parameters \( \gamma_m' \) and \( \gamma_d' \) are scaling factors, both empirically set to 1 in this study.
 
 \subsection{B. Construction of the Heterogeneous Graph}\label{graph}
 To comprehensively integrate the intricate relationships among oligopeptides, microbes, and diseases, a ternary heterogeneous graph is constructed, incorporating both structural and semantic enhancements. Based on the known inter-entity associations, the graph incorporates both structural edges and semantic edges derived from similarity matrices, thereby enhancing the representational richness and overall topological connectivity. Let \( S_p \), \( S_m \) and \( S_d \) denote the similarity matrices for oligopeptides, microbes, and diseases, respectively, and \( A_pm \), \( A_pd \), \( A_md \) denote the known binary association matrices among these entities. The structural and semantic edges are concatenated by node type to form a unified heterogeneous adjacency matrix \( \mathbf{M} \in \mathbb{R}^{(n_p + n_m + n_d) \times (n_p + n_m + n_d)} \), where  \( n_p \), \( n_m \) and \( n_d \) represent the number of oligopeptide, microbe, and disease nodes, respectively.
 \begin{equation}
 \mathbf{M} =
 \begin{bmatrix}
 S_p & A_{pm} & A_{pd} \\
 A_{pm}^{T} & S_m & A_{md} \\
 A_{pd}^{T} & A_{md}^{T} & S_d
 \end{bmatrix}.
 \end{equation}
 \begin{figure}
     \centering
     \includegraphics[width=0.48\textwidth]{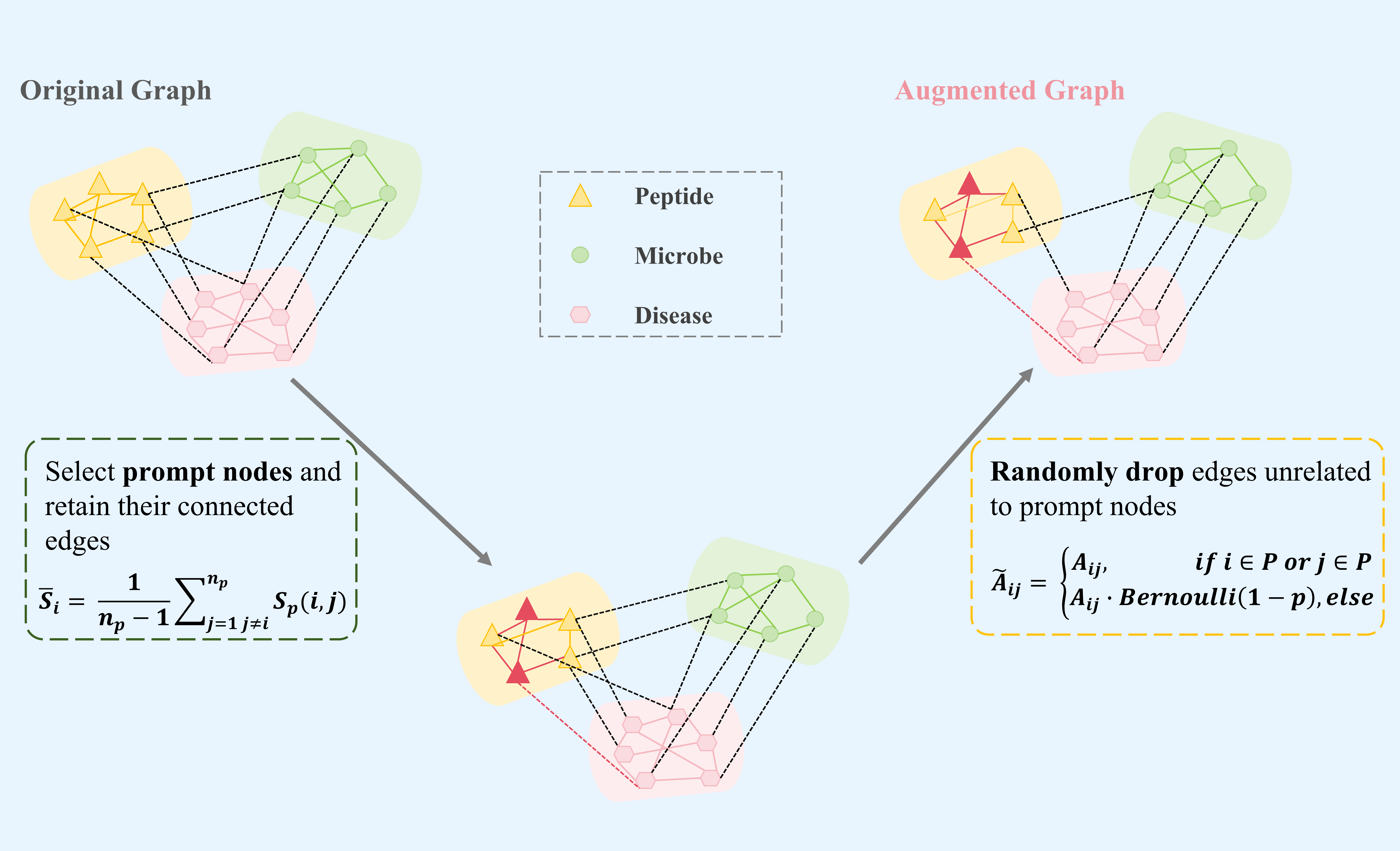}
     \caption{The prompt-guided graph augmentation strategy retains the edges of selected informative oligopeptide nodes while randomly perturbing those connected to non-prompt nodes, thereby preserving key structural regions and improving the quality of contrastive embeddings.}
     \label{fig:prompt}
 \end{figure}
 
 In this formulation, \( S_p \in \mathbb{R}^{n_p \times n_p} \) denotes the oligopeptide sequence similarity matrix computed using the Smith–Waterman alignment algorithm. \( S_m \in \mathbb{R}^{n_m \times n_m} \) and \( S_d \in \mathbb{R}^{n_d \times n_d} \) are the microbe–microbe and disease–disease similarity matrices, respectively, calculated via the Gaussian Interaction Profile (GIP) kernel. The binary association matrices \( A_{pm} \in \{0,1\}^{n_p \times n_m} \), \( A_{pd} \in \{0,1\}^{n_p \times n_d} \), and \( A_{md} \in \{0,1\}^{n_m \times n_d} \) encode the known links between oligopeptides and microbes, oligopeptides and diseases, and microbes and diseases, respectively.

 \subsection{C. Embedding Representation Learning}\label{representation}
 Following the construction of the heterogeneous graph, the model proceeds to the node embedding learning stage. Conventional graph augmentation approaches generally rely on randomly perturbing the adjacency matrix to create augmented views for contrastive learning. However, when applied to ternary heterogeneous graphs—where oligopeptides often serve as central hubs—such indiscriminate perturbations may disrupt structurally critical regions. This can compromise the model's ability to capture meaningful topological patterns and weaken the discriminative power of the learned embeddings. To mitigate this issue, a prompt-guided graph augmentation strategy is introduced (illustrated in Figure~\ref{fig:prompt}), which selectively preserves structurally informative regions while enabling contrastive view generation.
 
 The proposed augmentation strategy operates by first identifying structurally significant oligopeptide nodes—referred to as prompt nodes—based on their similarity to other peptides. For these prompt nodes, all connected edges are preserved. In contrast, edges linked to non-prompt nodes are subjected to stochastic perturbations to generate diversified graph views for contrastive learning.Prompt node selection is performed using the oligopeptide similarity matrix \( S_p \in \mathbb{R}^{n_p \times n_p} \), where the average similarity score of each oligopeptide \(i\) is calculated as:
 \begin{equation}
 \bar{s}_i = \frac{1}{n_p - 1} \sum_{\substack{j=1 \\ j \ne i}}^{n_p} S_p(i, j).
 \end{equation}

An oligopeptide is designated as a prompt node if its average similarity exceeds a predefined threshold \(\tau\):
  \begin{equation}
 \bar{s}_i > \tau.
 \end{equation}
 
 In this case, all edges connected to node i are preserved. Otherwise, it is treated as a non-prompt node and its adjacent edges are randomly dropped with a given probability. The edge perturbation process is formally defined as:
 \begin{equation}
 \tilde{A}_{ij} =
 \begin{cases}
 A_{ij}, & \text{if } i \in P \text{ or } j \in P \\
 A_{ij} \cdot \text{Bernoulli}(1 - p), & \text{else}
 \end{cases}.
 \end{equation}
 
 Herein, \(  A_{ij} \) denotes the edge weight between nodes \(i\) and \(j\) in the original graph, \(P\) denotes the set of prompt nodes, and \(p\) is the edge drop rate.

\begin{figure}
     \centering
     \includegraphics[width=0.48\textwidth]{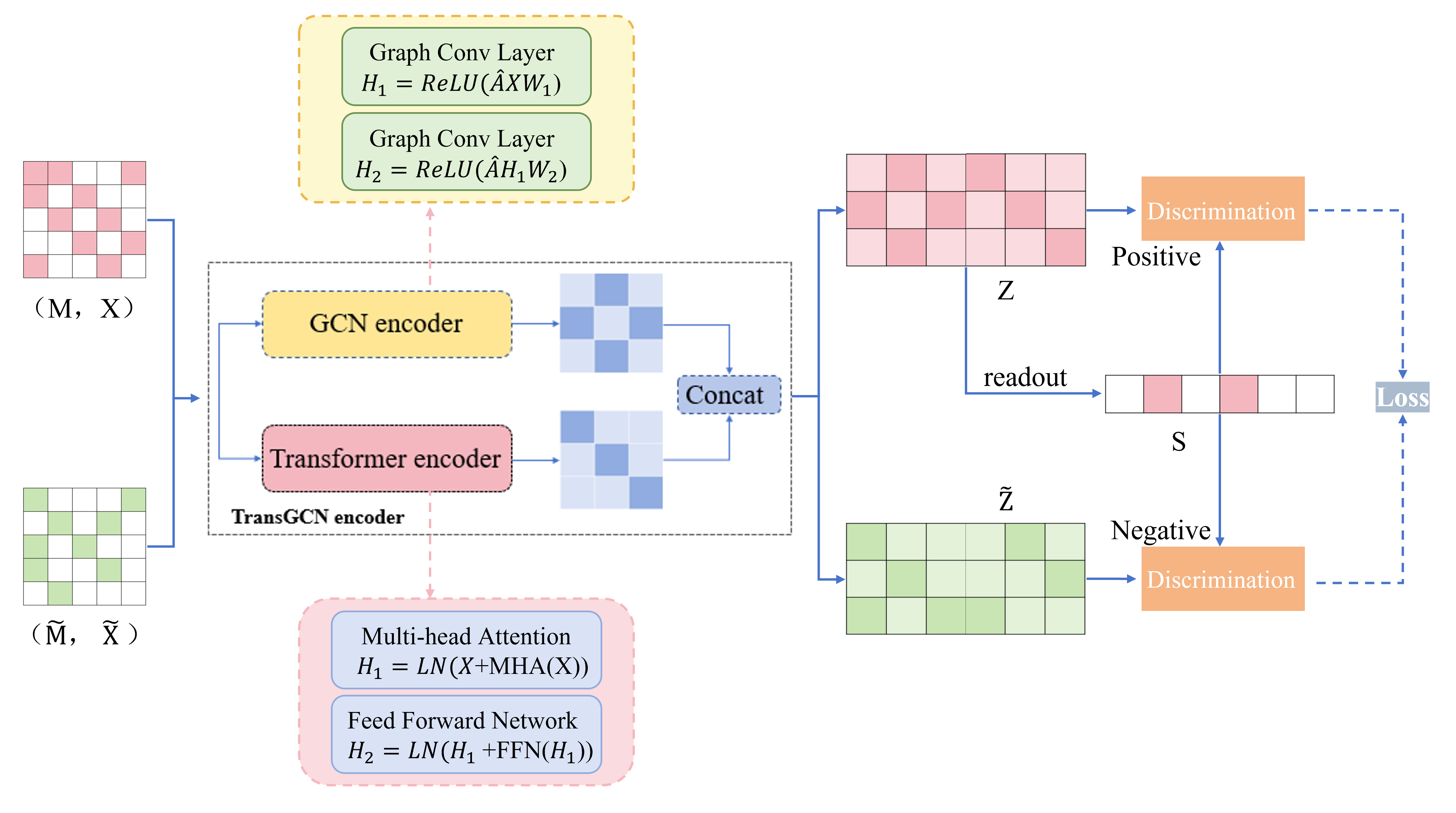}
     \caption{Illustration of the embedding learning module. Both the original and augmented graphs are fed into a dual-encoder architecture composed of a GCN and a Transformer to extract local structural features and global semantic dependencies, respectively. The resulting node embeddings are concatenated to form positive and negative pairs for contrastive learning. The learned representations are optimized through a contrastive loss to improve their discriminative power.}
     \label{fig:embedding}
 \end{figure}

 Once the original and augmented graphs are constructed, the model proceeds to the embedding learning phase (illustrated in Figure~\ref{fig:embedding}). A dual-encoder architecture is adopted, consisting of a Graph Convolutional Network (GCN) and a Transformer module. The GCN encoder is responsible for capturing local topological structures by aggregating neighborhood information, while also encoding intra-type semantic similarities derived from the constructed similarity matrices. In parallel, the Transformer encoder captures long-range dependencies and latent interactions between non-adjacent nodes using a self-attention mechanism. For instance, specific microbial nodes may serve as hubs connecting multiple oligopeptide–disease pairs, forming information-rich regions that facilitate semantic propagation across the graph. The synergy between GCN and Transformer encoders enables the model to effectively capture both fine-grained local structures and global semantic coherence, thereby enhancing the expressiveness and robustness of the learned node embeddings.
 
 For the original graph \( G \) and its augmented graph \( G' \), node embeddings are independently obtained by feeding them into the Graph Convolutional Network (GCN) encoder and the Transformer encoder. To effectively integrate local structural information and global semantic dependencies, the resulting embeddings from both encoders are concatenated to form the unified node representation:
 \begin{equation}
 \mathbf{z}_i = \left[ \mathbf{z}_i^{\mathrm{GCN}} \, \| \, \mathbf{z}_i^{\mathrm{Trans}} \right],
 \end{equation}
in Eq. (11), \( \mathbf{z}_i^{\mathrm{GCN}} \) and \( \mathbf{z}_i^{\mathrm{Trans}} \) represent the embeddings of node \( i \) generated by the GCN and Transformer encoders, respectively. Similarly, for the augmented graph \( G' \), the embedding \( \tilde{\mathbf{z}}_i \) is derived using the same dual-encoder architecture. To incorporate global contextual information, a graph-level embedding \( \mathbf{z}_g \) is computed by averaging all node embeddings in the original graph:
 \begin{equation}
 \mathbf{z}_g = \frac{1}{|\mathcal{V}|} \sum_{i \in \mathcal{V}} \mathbf{z}_i,
 \end{equation}
 where \( \mathcal{V} \) denotes the set of all nodes in the graph. This graph-level embedding is concatenated with each node embedding to form positive and negative samples for contrastive discrimination.

 Specifically, positive samples are created by concatenating \( \mathbf{z}_i \) (from the original graph) with \( \mathbf{z}_g \), while negative samples are formed by combining \( \tilde{\mathbf{z}}_i \) (from the augmented graph) with the same global vector. These concatenated vectors are then fed into a discriminator \( D(\cdot) \) and optimized using the binary cross-entropy loss function:
 \begin{equation}
 \mathcal{L}_{\text{contrast}} = - \log D([\mathbf{z}_i, \mathbf{z}_g]) - \log(1 - D([\tilde{\mathbf{z}}_i, \mathbf{z}_g])).
 \end{equation}

 Herein, \( [\mathbf{z}_i, \mathbf{z}_g] \) and \( [\tilde{\mathbf{z}}_i, \mathbf{z}_g] \) represent the concatenated representations of the positive and negative pairs, respectively.The discriminator is trained to distinguish between them, thereby guiding the encoders to bring positive pairs closer and push negative pairs apart in the latent embedding space.

 \subsection{D. Association Prediction Layer and loss Function}\label{prediction}
 Following the contrastive optimization of node representations, an association prediction module is constructed to infer potential links between oligopeptides and infectious diseases. The vector embeddings of oligopeptide and disease nodes are extracted from the learned representation space and subsequently used as input features for prediction. Specifically, the embeddings of oligopeptide node \({z}_p\) and disease node \({z}_d\) are concatenated to form the pairwise representation:
 \begin{equation}
 \mathbf{h}_{pd} = \left[ \mathbf{z}_p \, \| \, \mathbf{z}_d \right].
 \end{equation}

 The resulting concatenated vector \( \mathbf{h}_{pd} \) is subsequently passed through a multi-layer perceptron (MLP) to compute the predicted association score:
 \begin{equation}
 \hat{y}_{pd} = \mathrm{MLP}(\mathbf{h}_{pd}).
 \end{equation}

 Here, \( \hat{y}_{pd} \in (0,1) \) represents the predicted probability of an existing association between the given oligopeptide–disease pair. The MLP comprises multiple nonlinear fully connected layers, with a Sigmoid activation function applied in the final layer to ensure the output is bounded within the interval (0, 1). To enable supervised training, the ground truth label \( y_{pd} \in \{0, 1\} \) is used as the supervision signal for optimizing the association prediction task. Specifically, \( y_{pd} = 1 \) indicates a known association, whereas \( y_{pd} = 0 \) denotes the absence of such a relationship. The discrepancy between the predicted score \( \hat{y}_{pd} \) and the true label \( y_{pd} \) is minimized using the binary cross-entropy loss function, defined as:
 \begin{equation}
 \mathcal{L}_{\mathrm{pred}} = - y_{pd} \log \hat{y}_{pd} - (1 - y_{pd}) \log (1 - \hat{y}_{pd}).
 \end{equation}

 The final training objective integrates the contrastive loss \( \mathcal{L}_{\mathrm{contrast}} \) with the prediction loss \( \mathcal{L}_{\mathrm{pred}} \) as follows:
 \begin{equation}
 \mathcal{L} = \mathcal{L}_{\mathrm{contrast}} + \lambda \mathcal{L}_{\mathrm{pred}},
 \end{equation}
 where \( \lambda \) is a tunable hyperparameter that balances the contributions of the two loss components during training. This joint optimization framework effectively enhances the discriminative capability of structural embeddings and improves the accuracy of oligopeptide–disease association inference.
 
\section{IV Experimental Results}\label{Experiment}
 \subsection{A. Experimental Settings}\label{exset}
 
 To systematically evaluate the performance of the model in predicting associations between oligopeptides and infectious diseases, extensive experiments are conducted on the constructed ternary heterogeneous graph consisting of oligopeptides, microbes, and diseases. All experiments are conducted under a consistent hardware and software environment, with the model trained and evaluated using five-fold cross-validation. This process is repeated five times, and the average performance across the five runs is reported to enhance evaluation stability and reliability. To assess classification performance, several standard binary classification metrics are employed, including Accuracy, Precision, Recall, F1-score, AUC, and AUPR. AUC and AUPR, representing the areas under the ROC and Precision–Recall curves, respectively, are particularly suitable for evaluating performance on imbalanced datasets. The calculation formulas are as follows:
 \begin{align}
 \mathrm{TPR} &= \frac{\mathrm{TP}}{\mathrm{TP} + \mathrm{FN}},  \\
 \mathrm{FPR} &= \frac{\mathrm{FP}}{\mathrm{FP} + \mathrm{TN}}, \\
 \mathrm{Precision} &= \frac{\mathrm{TP}}{\mathrm{TP} + \mathrm{FP}},  \\
 \mathrm{Recall} &= \frac{\mathrm{TP}}{\mathrm{TP} + \mathrm{FN}},  \\
 \mathrm{Accuracy} &= \frac{\mathrm{TP} + \mathrm{TN}}{\mathrm{TP} + \mathrm{TN} + \mathrm{FP} + \mathrm{FN}},  \\
 \mathrm{F1\mbox{-}score} &= \frac{2 \times \mathrm{TP}}{2 \times \mathrm{TP} + \mathrm{FP} + \mathrm{FN}}.
 \end{align}

 In the above formulas, TP (True Positive) denotes the number of samples correctly predicted as positive, and TN (True Negative) denotes those correctly predicted as negative. FP (False Positive) and FN (False Negative) represent the numbers of samples incorrectly predicted as positive and negative, respectively. Precision measures the proportion of true positives among all predicted positives, while Recall indicates the proportion of actual positives correctly identified. The F1-score, defined as the harmonic mean of Precision and Recall, is used to evaluate model robustness under class imbalance.

 \begin{figure*}[ht]
  \centering
  \includegraphics[width=\textwidth]{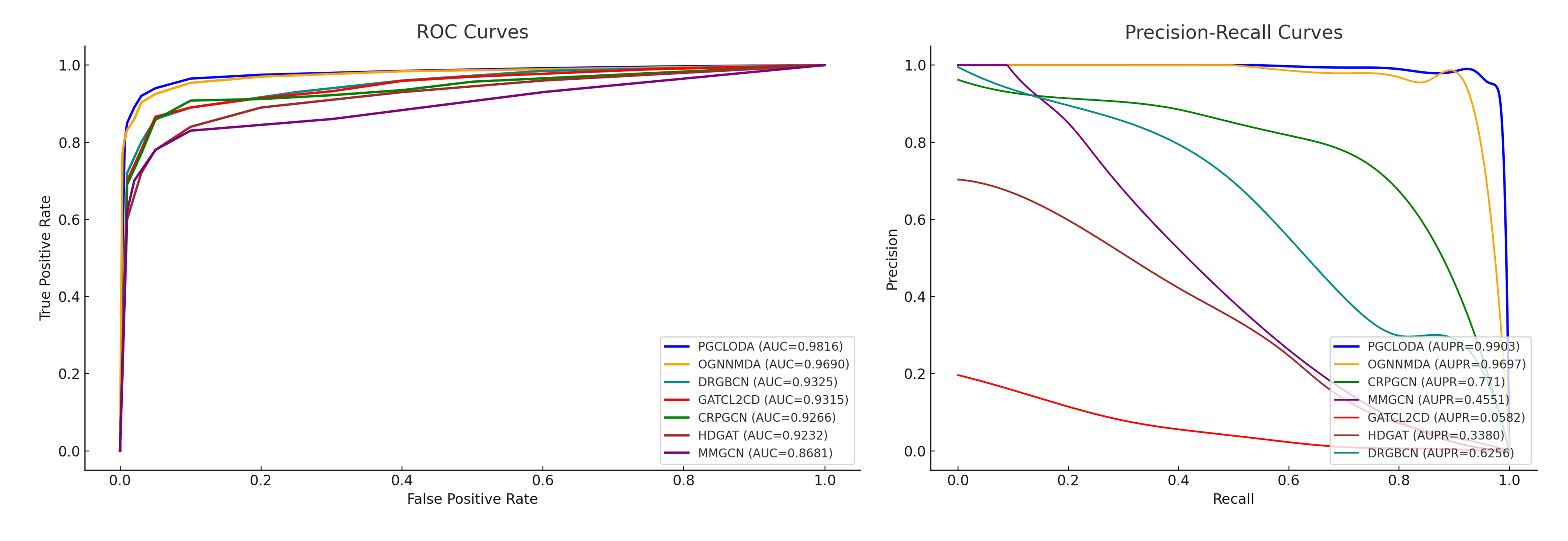}
  \caption{The left panel presents the ROC curves for all models on the oligopeptide–disease association task, where the x-axis denotes the False Positive Rate and the y-axis denotes the True Positive Rate. The right panel displays the Precision–Recall (PR) curves, where Recall is plotted on the x-axis and Precision on the y-axis. The proposed method consistently outperforms all baseline models on both metrics, demonstrating superior discriminative capability and robustness.}
  \label{fig:ROCPR}
 \end{figure*}

 \subsection{B. Comparative Experiments}\label{comex}
 To evaluate the effectiveness of the proposed framework for oligopeptide–infectious disease association prediction, six state-of-the-art graph-based association prediction models are selected for comparison. The comparative models include MMGCN, GATCL2CD, OGNNMDA, CRPGCN, HDGAT, and DRGBCN. These models have been widely applied to various biological association prediction tasks, including miRNA–disease, circRNA–disease, and drug–disease prediction, reflecting recent advances in multi-view learning, attention mechanisms, and contrastive learning. The details of each comparison model are as follows:
 
 \begin{table*}[htbp]
  \centering
  \caption{Performance comparison of different models across six metrics.}
  \label{tab:comparison}
  \renewcommand{\arraystretch}{1.2}
  \begin{tabular}{l c c c c c c}
    \toprule
    \textbf{Module} & \textbf{AUROC} & \textbf{AUPRC} & \textbf{F1} & \textbf{Accuracy} & \textbf{Recall} & \textbf{Precision} \\
    \midrule
    MMGCN     & 0.8681 & 0.4551 & 0.9175 & 0.9174 & 0.2517 & 0.5807 \\
    GATCL2CD  & 0.9315 & 0.0582 & 0.0171 & 0.4967 & 0.9321 & 0.0093 \\
    OGNNMDA   & 0.9690 & 0.9697 & 0.9174 & 0.9172 & 0.9191 & 0.9156 \\
    CRPGCN    & 0.9266 & 0.7710 & 0.5238 & 0.9311 & 0.4622 & 0.8980 \\
    HDGAT     & 0.9232 & 0.3380 & 0.3552 & 0.9193 & 0.6623 & 0.2441 \\
    DRGBCN    & 0.9325 & 0.6256 & 0.3723 & 0.7520 & 0.9418 & 0.2320 \\
    \textbf{Ours} & \textbf{0.9816} & \textbf{0.9903} & \textbf{0.9525} & \textbf{0.9370} & \textbf{0.9602} & \textbf{0.9450} \\
    \bottomrule
  \end{tabular}
 \end{table*}

 \textbf{MMGCN:} Enhances node feature representations by integrating multiple similarity views using multi-channel GCNs, but lacks the capability to model path-level semantics in heterogeneous graphs. 
 
 \textbf{GATCL2CD:} Integrates graph attention and contrastive learning to enhance the structural discriminability of node embeddings, but is primarily designed for homogeneous graph scenarios.
 
 \textbf{OGNNMDA:} Alleviates over-smoothing by applying graph perturbations and contrastive learning to promote feature diversity, yet does not explicitly model heterogeneous entity types.
 
 \textbf{CRPGCN:} Constructs structural graphs using random walks and attribute features, and aggregates node embeddings through GCNs, but lacks the capacity to model semantic dependencies across different entity types.
 
 \textbf{HDGAT:} Captures node importance through both local and global attention mechanisms, but is designed for static graphs and lacks modeling of semantic roles among multiple node types.
 
 \textbf{DRGBCN:} Models structural semantics using a hybrid of Transformer and multi-layer GCN, and aligns multi-view representations via contrastive loss, but does not incorporate heterogeneous prompt-based augmentation.

 All models are trained using identical data splits and hyperparameter configurations, and evaluated under five-fold cross-validation. Performance comparisons are made across six metrics—AUROC, AUPRC, F1-score, Accuracy, Recall, and Precision—as summarized in Table~\ref{tab:comparison}.
    \begin{figure}
     \centering
     \includegraphics[width=0.48\textwidth]{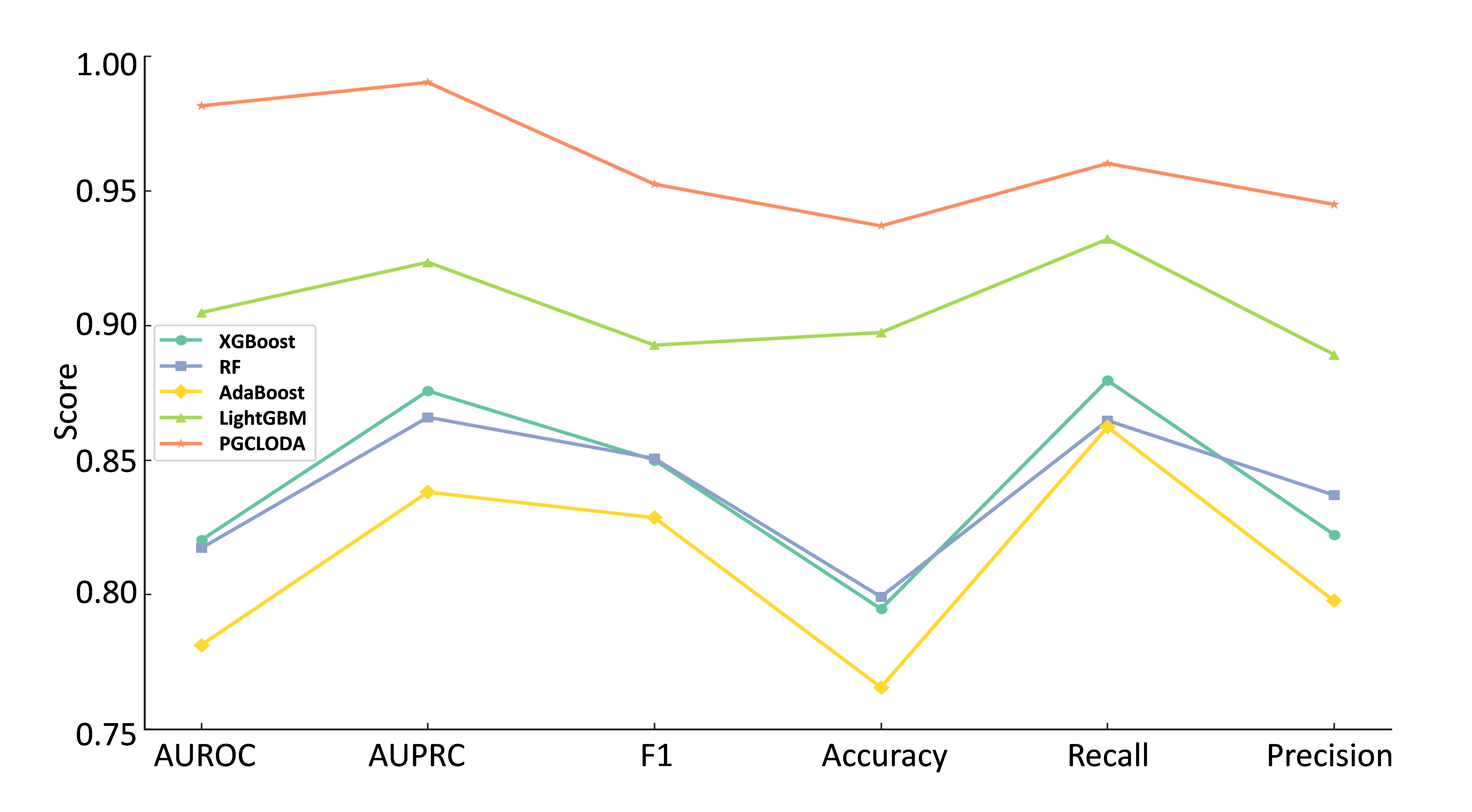}
     \caption{Performance comparison across six evaluation metrics (AUROC, AUPRC, F1-score, Accuracy, Recall, and Precision) when the MLP predictor is replaced by alternative classifiers. The label "Ours" denotes the proposed model using the MLP predictor. The proposed MLP predictor achieves consistently superior performance across all metrics compared to alternative classifiers, including XGBoost, Random Forest (RF), AdaBoost, and LightGBM. This suggests that MLP is more effective at capturing complex nonlinear relationships among features, making it particularly suitable for association prediction tasks within complex heterogeneous structures such as oligopeptides and infectious diseases}
     \label{fig:MLP}
  \end{figure}
 The proposed framework outperforms all baselines across all evaluation metrics. Notably, substantial gains in AUPRC and F1-score underscore the framework’s effectiveness in identifying associations under class imbalance and structurally complex conditions. Additionally, the ROC and PR curves of all models are presented in Figure~\ref{fig:ROCPR} to illustrate stability and generalization performance under varying classification thresholds. As illustrated, the proposed framework exhibits steeper ROC and PR curves with larger areas under the curves, confirming its superior predictive capability.
 
 \subsection{C. Ablation Experiments}\label{able}
   \begin{figure*}[htbp]
   \centering
   \includegraphics[width=\textwidth]{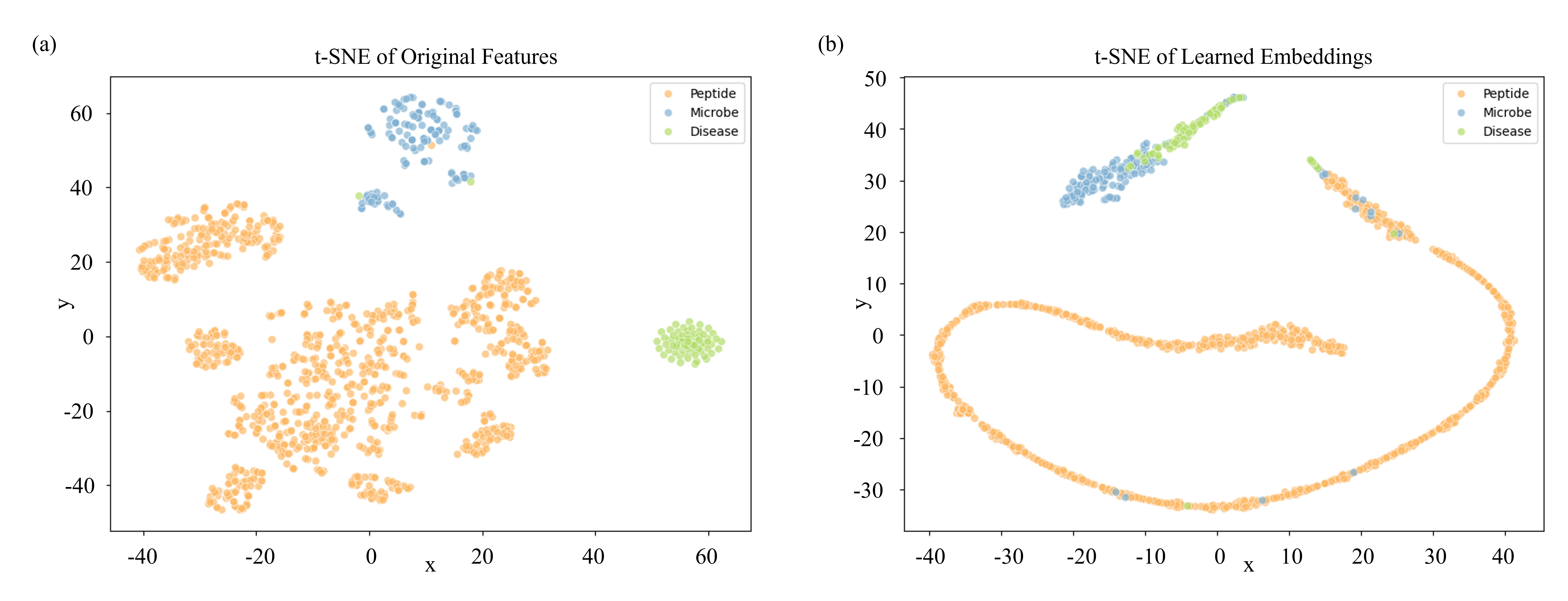}
   \caption{Comparison of t-SNE visualizations before and after encoding. The learned embeddings demonstrate more compact clusters and clearer separation among peptides, microbes, and diseases, supporting the effectiveness of the proposed contrastive learning framework.}
   \label{fig:t-SNE}
  \end{figure*} 
 To assess the contribution of individual components within the proposed model, four ablation experiments were conducted, targeting the predictor structure, the contrastive learning mechanism, the inclusion of microbe nodes, and the encoder architecture.

 First, to evaluate the effectiveness of the multilayer perceptron (MLP) as the final prediction module, we replaced it with four widely used machine learning classifiers: XGBoost, Random Forest (RF), AdaBoost, and LightGBM. Comparative results across six metrics are presented in Figure~\ref{fig:MLP}. The results demonstrate that the proposed model consistently outperforms all alternative classifiers across all six evaluation metrics, including AUROC, AUPRC, F1-score, Accuracy, Recall, and Precision. This indicates that the MLP, serving as a nonlinear discriminative module, possesses superior capability in multi-source feature integration and is particularly effective for association prediction in complex heterogeneous graphs.

 \begin{table*}[htbp]
  \centering
  \caption{Ablation results of key modules in the proposed model.}
  \label{tab:ablation}
  \renewcommand{\arraystretch}{1.2}
  \begin{tabular}{|l|c|c|c|c|c|c|}
    \hline
    \textbf{Ablation configurations} & \textbf{AUROC} & \textbf{AUPRC} & \textbf{F1} & \textbf{Accuracy} & \textbf{Recall} & \textbf{Precision} \\
    \hline
    \multicolumn{7}{|l|}{\textbf{Contrastive Learning Module}} \\
    \hline
    w/o Contrast & 0.8307 & 0.8821 & 0.8097 & 0.6915 & 0.8947 & 0.6813 \\
    Full model   & 0.9816 & 0.9903 & 0.9525 & 0.9370 & 0.9602 & 0.9450 \\
    \hline
    \multicolumn{7}{|l|}{\textbf{Microbe Node Module}} \\
    \hline
    w/o Microbe  & 0.9498 & 0.9715 & 0.9180 & 0.8895 & 0.9408 & 0.8963 \\
    Full model   & 0.9816 & 0.9903 & 0.9525 & 0.9370 & 0.9602 & 0.9450 \\
    \hline
    \multicolumn{7}{|l|}{\textbf{Encoder Module}} \\
    \hline
    w/o GCN         & 0.9014 & 0.9705 & 0.9493 & 0.9110 & 0.9615 & 0.9238 \\
    w/o Transformer & 0.8088 & 0.8694 & 0.8102 & 0.7339 & 0.8771 & 0.7528 \\
    Full model      & 0.9816 & 0.9903 & 0.9525 & 0.9370 & 0.9602 & 0.9450 \\
    \hline
  \end{tabular}
 \end{table*}
 
 Furthermore, to comprehensively evaluate the impact of core components, we performed ablation analyses by individually removing the contrastive learning module, microbial nodes, and either the GCN or Transformer from the encoder. The resulting changes in six evaluation metrics are summarized in Table 3. Experimental results show that removing the contrastive learning module significantly degrades model performance, with AUPRC and F1-score dropping to 0.8821 and 0.8097, respectively. This demonstrates that structural contrastive learning substantially improves the consistency and discriminative power of the learned embeddings.When microbial nodes were removed—leaving only binary relationships between oligopeptides and diseases—all evaluation metrics declined to varying degrees. This validates the semantic bridging role of microbes in the ternary structure, especially in modeling information propagation for infectious diseases. With the GCN module removed from the encoder, model performance slightly declined but still maintained reasonable accuracy. In contrast, removing the Transformer led to a more substantial performance drop, highlighting the critical role of global dependency modeling in capturing complex path semantics and cross-type interactions.

 These results highlight the essential contributions of each module within the proposed model, particularly under the ternary heterogeneous graph structure. In particular, contrastive learning and the dual-encoder architecture play key roles in enhancing embedding quality and capturing global semantic dependencies.

 To further demonstrate the representational advantages conferred by the proposed dual-encoder and contrastive learning framework, we perform t-SNE visualizations on both the original input features and the learned embeddings, as shown in Figure~\ref{fig:t-SNE}. In Figure 6(a), the original features exhibit significant overlap among peptides, microbes, and diseases, with poorly separated and highly entangled distributions. In contrast, Figure 6(b) illustrates that the embeddings produced by our model exhibit enhanced intra-class compactness and inter-class separability. This visual evidence clearly demonstrates that incorporating contrastive embedding mechanisms significantly enhances the discriminative capacity and informativeness of node representations, thereby providing a more solid foundation for downstream association prediction tasks.

 \subsection{D. Hyperparameter Experiments}\label{hypere}
\begin{figure}
     \centering
     \includegraphics[width=0.48\textwidth]{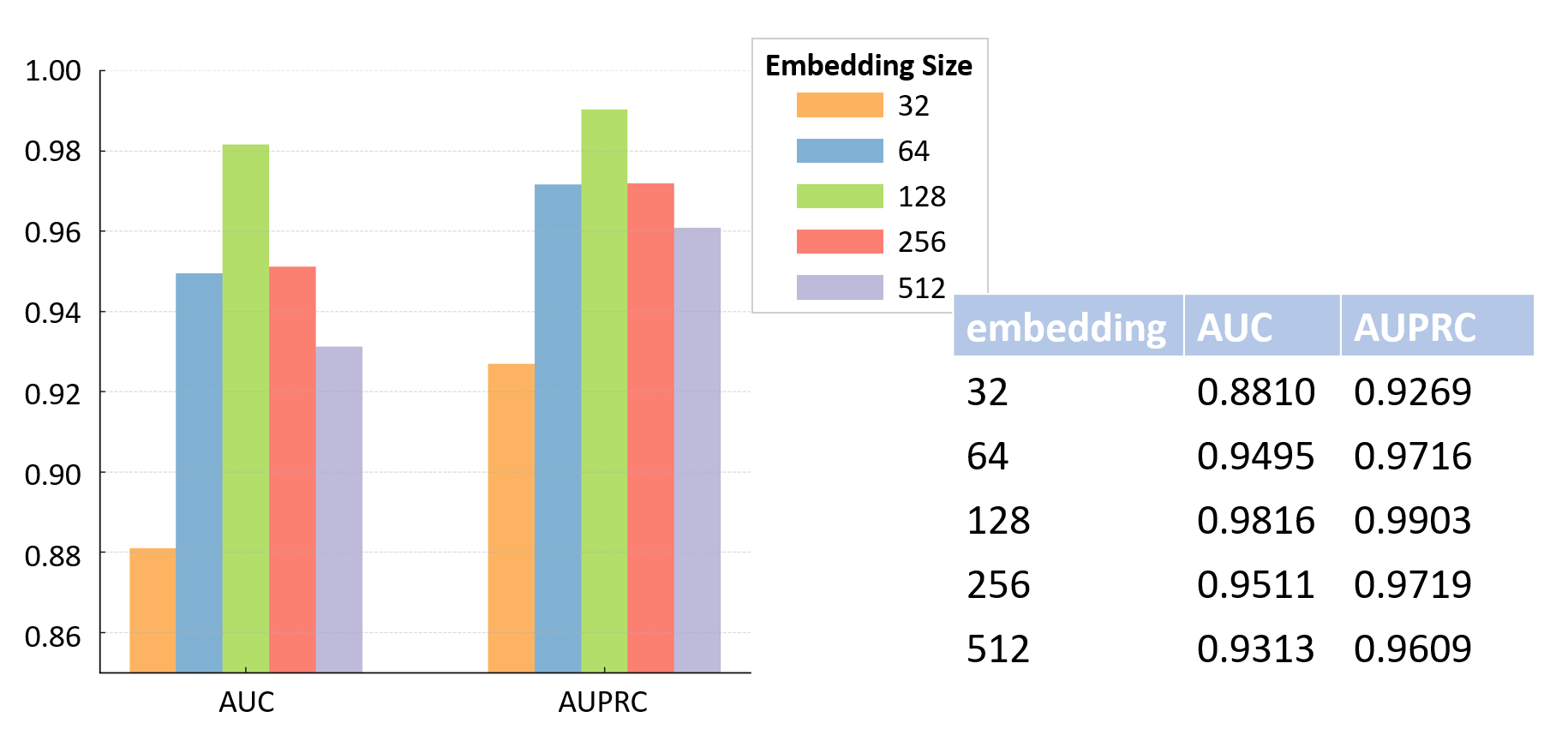}
     \caption{Peak AUROC and AUPRC are achieved at an embedding dimension of 128, with moderate performance degradation observed at both smaller (e.g., 32) and larger (e.g., 256, 512) dimensions.}
     \label{fig:embed}
 \end{figure}

\begin{figure}
     \centering
     \includegraphics[width=0.48\textwidth]{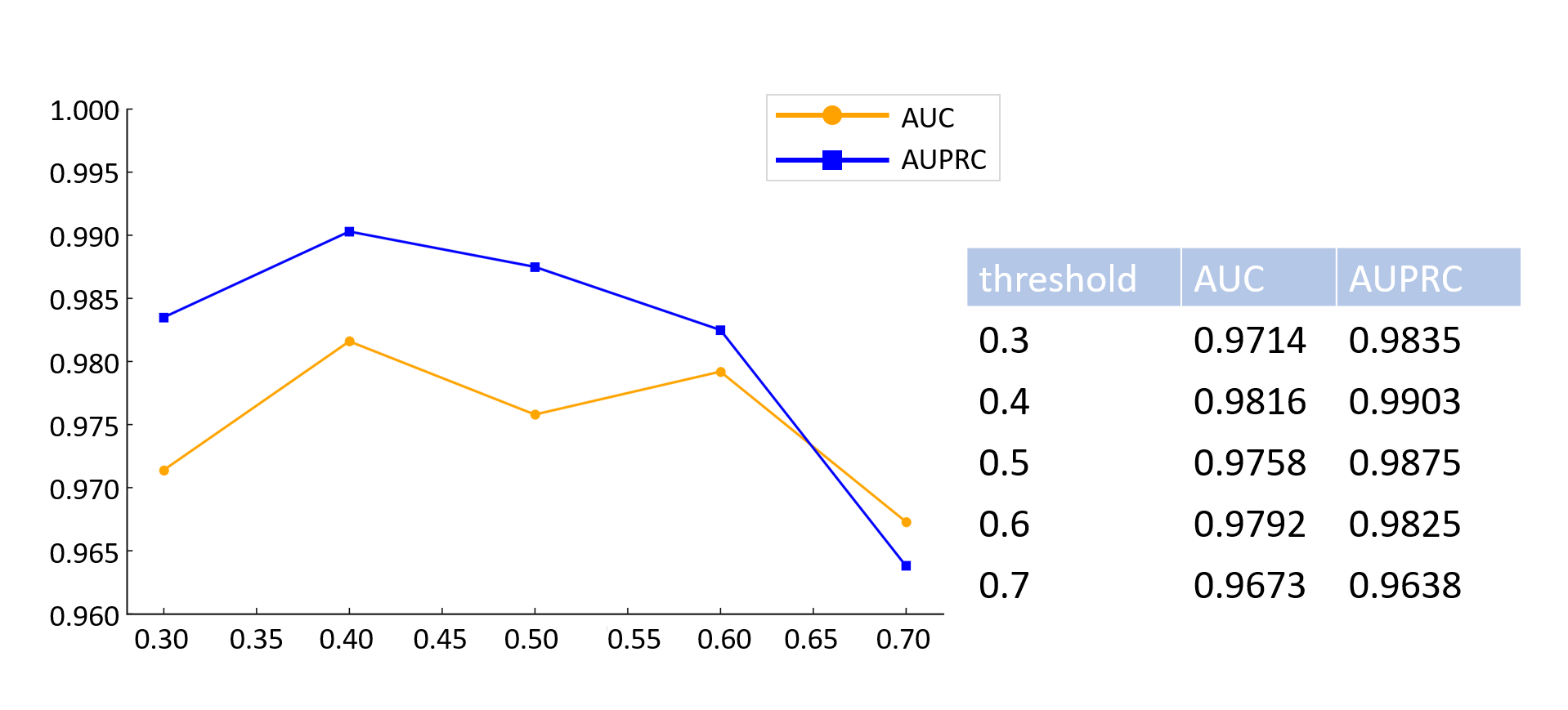}
     \caption{Peak performance is observed at a threshold of 0.4 (AUROC = 0.9816, AUPRC = 0.9903), with slight performance degradation at both lower and higher thresholds.}
     \label{fig:threshold}
 \end{figure}

 To investigate the impact of critical hyperparameters on model performance, sensitivity analyses were conducted on the embedding dimension and the anchor node selection threshold. In each experiment, all other hyperparameters were held constant while varying only the target parameter. The corresponding AUROC and AUPRC trends are illustrated in Figure~\ref{fig:embed} and Figure~\ref{fig:threshold}.

 In the embedding dimension experiment, five values (32, 64, 128, 256, and 512) were evaluated. The model achieved optimal performance at an embedding size of 128, yielding an AUROC of 0.9816 and an AUPRC of 0.9903. Smaller embedding dimensions constrain representational capacity, while excessively large dimensions may result in overfitting and reduced generalization ability. Thus, selecting an appropriate embedding dimension is crucial for capturing the complex structural semantics of the ternary heterogeneous graph, while ensuring training stability and high predictive accuracy.

 Five anchor-node selection thresholds ranging from 0.3 to 0.7 were evaluated. PGCLODA achieved optimal performance at a threshold of 0.4, with an AUROC of 0.9816 and an AUPRC of 0.9903, significantly outperforming other settings. These findings underscore the importance of anchor-node selection in determining the effectiveness of graph augmentation. An appropriate threshold facilitates the selection of semantically representative nodes, preserves structural integrity during augmentation, and enhances the model’s structural discrimination in contrastive learning. In contrast, a low threshold may introduce excessive redundant nodes and insufficient perturbations, while a high threshold may eliminate informative structures, thereby reducing the consistency between the augmented and original graphs and impairing the performance of contrastive learning.

  \begin{table}[htbp]
  \setlength{\tabcolsep}{0.5pt} 
  \centering
  \caption{Performance of PGCLODA under varying positive-to-negative sample ratios.}
  \label{tab:imbalance}
  \renewcommand{\arraystretch}{1.25}
  \begin{tabular}{ccccccc}
  \toprule
  \textbf{Ratio} & \textbf{AUROC} & \textbf{AUPRC} & \textbf{F1} & \textbf{Accuracy} & \textbf{Recall} & \textbf{Precision} \\
  \midrule
  1:10 & 0.8991 & 0.6808 & 0.5922 & 0.9248 & 0.4409 & 0.9034 \\
  1:5 & 0.9147 & 0.7822 & 0.6834 & 0.9129 & 0.5643 & 0.8665 \\
  1:2 & 0.9302 & 0.8910 & 0.7691 & 0.8614 & 0.6924 & 0.8650 \\
  \textbf{1:1} & \textbf{0.9816} & \textbf{0.9903} & \textbf{0.9525} & \textbf{0.9370} & \textbf{0.9602} & \textbf{0.9450} \\
  \bottomrule
  \end{tabular}
  \end{table}
 \begin{table*}[htbp]
 \centering
 \caption{Top 10 predicted oligopeptide–microbe–disease triplets with literature support.}
 \label{tab:case-study}
 \renewcommand{\arraystretch}{1.3}
 \begin{tabularx}{\textwidth}{|c|X|X|X|X|X|X|}
 \hline
 \textbf{Rank} & \textbf{peptide name} & \textbf{microbe name} & \textbf{Evidence} & \textbf{microbe name} & \textbf{disease name} & \textbf{Evidence} \\
 \hline
 1 & RWRWRWRW & \textit{Fusarium solani} & PMID: 23203110 & \textit{Fusarium solani} & Keratitis & PMID: 32134799 \\
 \hline
 2 & FRIRVRV & \textit{Pseudomonas aeruginosa} & PMID: 28178190 & \textit{Pseudomonas aeruginosa} & Chronic lung disease & PMID: 39015565 \\
 \hline
 3 & XSYNGNSN & \textit{Staphylococcus aureus} & Unconfirmed & \textit{Staphylococcus aureus} & Acute skin abscess & Unconfirmed \\
 \hline
 4 & KIGAKI & \textit{Escherichia coli} & PMID: 11352918 & \textit{Escherichia coli} & Gastroenteritis & PMID: 31036328 \\
 \hline
 5 & FRIRVRV & \textit{Staphylococcus aureus} & PMID: 28178190 & \textit{Staphylococcus aureus} & Periodontal disease & PMID: 37770865 \\
 \hline
 6 & YTRGLPM & \textit{Staphylococcus aureus} & Unconfirmed & \textit{Staphylococcus aureus} & Acute skin abscess & PMID: 33303329 \\
 \hline
 7 & DEDLDE & \textit{Staphylococcus aureus} & PMID: 28299865 & \textit{Staphylococcus aureus} & Acute skin abscess & PMID: 33303329 \\
 \hline
 8 & RKKFWF & \textit{Penicillium expansum} & PMID: 11976121 & \textit{Penicillium expansum} & Caries & Unconfirmed \\
 \hline
 9 & KVFLGLK & \textit{Streptococcus pneumoniae} & PMID: 21268582 & \textit{Streptococcus pneumoniae} & Pneumonia & PMID: 28735461 \\
 \hline
 10 & DEKGPKWKR & \textit{Candida albicans} & PMID: 17272268 & \textit{Candida albicans} & Bacterial vaginosis & PMID: 25775428 \\
 \hline
 \end{tabularx}
 \end{table*}

 \subsection{E. Imbalance Robustness Experiment}\label{IRe}
  To evaluate the robustness of PGCLODA under varying degrees of class imbalance, we conducted experiments with four positive-to-negative sample ratios: 1:1, 1:2, 1:5, and 1:10. For each ratio, we employed a five-fold cross-validation protocol using the same model architecture and hyperparameter settings. This design isolates the effect of sample imbalance on model performance while keeping all other variables fixed. Evaluation metrics included AUROC, AUPRC, F1-score, Accuracy, Recall, and Precision.

  As shown in Table~\ref{tab:imbalance}, PGCLODA maintains competitive performance across different levels of class imbalance. Notably, the AUROC remains above 0.89 and the AUPRC above 0.68 even under the extreme 1:10 imbalance setting. With increasing class balance, performance improves consistently, reaching peak scores of 0.9816 AUROC and 0.9903 AUPRC at the 1:1 ratio. Furthermore, while the recall score drops under high imbalance (0.4409 for 1:10), the precision remains above 0.86 in all scenarios, indicating the model's strong ability to avoid false positives even with limited positive samples.

  These results demonstrate that PGCLODA is resilient to skewed class distributions, highlighting its potential applicability in real-world biomedical tasks where positive associations are typically sparse.
  
 \subsection{F. Case Study}\label{casestudy}
  To further demonstrate the practical applicability of PGCLODA in identifying novel oligopeptide–infectious disease associations, a case study was conducted on the top ten oligopeptide–microbe–disease triplets with the highest prediction confidence among unlabeled samples outside the training set. The results, summarized in Table~\ref{tab:case-study}, list the biological entities involved in each predicted triplet, along with supporting literature evidence (when available), including PubMed identifiers for both the oligopeptide–microbe and microbe–disease associations.
 
 Among the top ten high-confidence predictions, several oligopeptide–microbe and microbe–disease associations have been previously documented in the literature. For instance, the peptide RWRWRWRW is predicted to associate with the fungus Fusarium solani, which has been implicated in keratitis-related studies (PMID: 23203110). Similarly, the predicted interaction between FRIRVRV and Pseudomonas aeruginosa is supported by PMID: 28178190, and its involvement in chronic lung disease has been validated in PMID: 39015565. Likewise, Staphylococcus aureus is associated with multiple peptides (e.g., XSYNGNSN, FRIRVRV, YTRGLPM, DEDLDE), underscoring its pivotal pathogenic role in acute skin abscesses, consistent with previously reported evidence (e.g., PMID: 33303329). Although some predicted associations remain unconfirmed in current public databases, their structural similarity and contextual relevance to verified pathways suggest considerable potential for future biological investigation. The predictions generated by PGCLODA not only align with known associations documented in existing knowledge bases but also uncover previously overlooked or unclassified triplets, thereby offering promising candidates for downstream biological validation. This case study demonstrates that PGCLODA exhibits strong generalization capability and novel association discovery potential within complex ternary heterogeneous graphs, thereby offering valuable data support for elucidating infectious disease mechanisms and advancing peptide-based drug discovery. 

\section{V Conclusion and Outlook}\label{conclusion}
This study addresses the challenge of uncovering potential associations between oligopeptides and infectious diseases by proposing a deep learning framework grounded in a heterogeneous graph that jointly models three types of biological entities—oligopeptides, microbes, and diseases—along with their multi-level interrelationships. During graph construction, inter-node biological associations were enriched by integrating disease semantic similarities, microbial genomic features, and oligopeptide sequence similarities.To facilitate embedding representation learning, a graph augmentation strategy guided by anchor-node selection was introduced, along with a dual-encoder architecture—comprising a Graph Convolutional Network (GCN) and a Transformer—to capture both local adjacency patterns and global semantic dependencies. TA contrastive learning objective was further incorporated to enhance embedding consistency and discriminative capability within the heterogeneous graph. Finally, the learned embeddings were fused to construct a high-precision prediction model for oligopeptide–disease association inference. Experimental results demonstrate that PGCLODA consistently outperforms state-of-the-art methods across multiple evaluation metrics, validating its effectiveness and generalization capability in complex heterogeneous graph scenarios.

Despite these promising results in both predictive performance and model architecture, several directions remain open for future investigation. First, the current approach primarily relies on structural similarity for node attribute representation. Future work could leverage large-scale protein language models to enable contextual semantic modeling of oligopeptide sequences, thereby enhancing representation fidelity. Second, the current framework models the graph as static and thus fails to capture the temporal evolution of oligopeptide–microbe–disease interactions during disease progression. Incorporating a dynamic graph modeling mechanism could potentially address this limitation. Additionally, external knowledge graphs and multimodal biological data have not yet been incorporated for semantic enrichment. Exploring cross-modal alignment strategies and knowledge-guided mechanisms may further improve both the interpretability and biological plausibility of the model’s outputs. Overall, PGCLODA presents a novel paradigm for modeling interactions among complex biological entities and offers a valuable reference for representation learning on multi-source heterogeneous graphs, with promising potential for broader generalization and application in biomedical research.

\vspace{1em} 
\noindent
\fbox{%
  \begin{minipage}{0.92\linewidth}
  \textbf{Key points}
  \begin{itemize}
    \item PGCLODA constructs a ternary heterogeneous graph integrating oligopeptides, microbes, and diseases, effectively modeling indirect semantic pathways for infectious disease prediction.
    \item A prompt-guided contrastive learning mechanism is introduced, where anchor nodes guide structural perturbation to generate informative augmented views, enhancing embedding discriminability.
    \item A dual-encoder architecture combining GCN and Transformer is designed to jointly capture local adjacency and global semantic dependencies across heterogeneous node types.
  \end{itemize}
  \end{minipage}
}

\bibliographystyle{unsrt}
\bibliography{arxiv}
\end{document}